\documentclass{bmvc2k}

\title{Single Image 3D Hand Reconstruction with Mesh Convolutions}

\addauthor{Dominik Kulon\textsuperscript{1, }}{d.kulon17@imperial.ac.uk}{3}
\addauthor{Haoyang Wang\textsuperscript{1, }}{haoyang.wang15@imperial.ac.uk}{3}
\addauthor{Riza Alp G\"uler\textsuperscript{1, }}{r.guler@imperial.ac.uk}{3}
\addauthor{Michael Bronstein\textsuperscript{1, 2, }}{m.bronstein@imperial.ac.uk}{4}
\addauthor{Stefanos Zafeiriou\textsuperscript{1, }}{s.zafeiriou@imperial.ac.uk}{3}

\addinstitution{
 Imperial College London \\
 Department of Computing  \\
 London, UK
}
\addinstitution{
Universit\`a della Svizzera italiana \\
Institute of Computational Science \\
Lugano, Switzerland
}

\addinstitution{
Ariel AI \\
arielai.com
}

\addinstitution{
Twitter, Inc. \\
twitter.com
}

\runninghead{Kulon et al.}{Single Image 3D Hand Reconstruction}

\begin{document}

\maketitle

\begin{abstract}

Monocular 3D reconstruction of deformable objects, such as human body parts, has been typically approached by predicting parameters of heavyweight linear models. In this paper, we demonstrate an alternative solution that is based on the idea of encoding images into a latent non-linear representation of meshes.
The prior on 3D hand shapes is learned by training an autoencoder with intrinsic graph convolutions performed in the spectral domain. 
The pre-trained decoder acts as a non-linear statistical deformable model. The latent parameters that reconstruct the shape and articulated pose of hands in the image are predicted using an image encoder.
We show that our system reconstructs plausible meshes and operates in real-time. We evaluate the quality of the mesh reconstructions produced by the decoder on a new dataset and show latent space interpolation results. 
Our code, data, and models will be made publicly available.
\end{abstract}

%-------------------------------------------------------------------------
\section{Introduction}

\begin{figure*}
\begin{center}
  \includegraphics[width=0.95\linewidth]{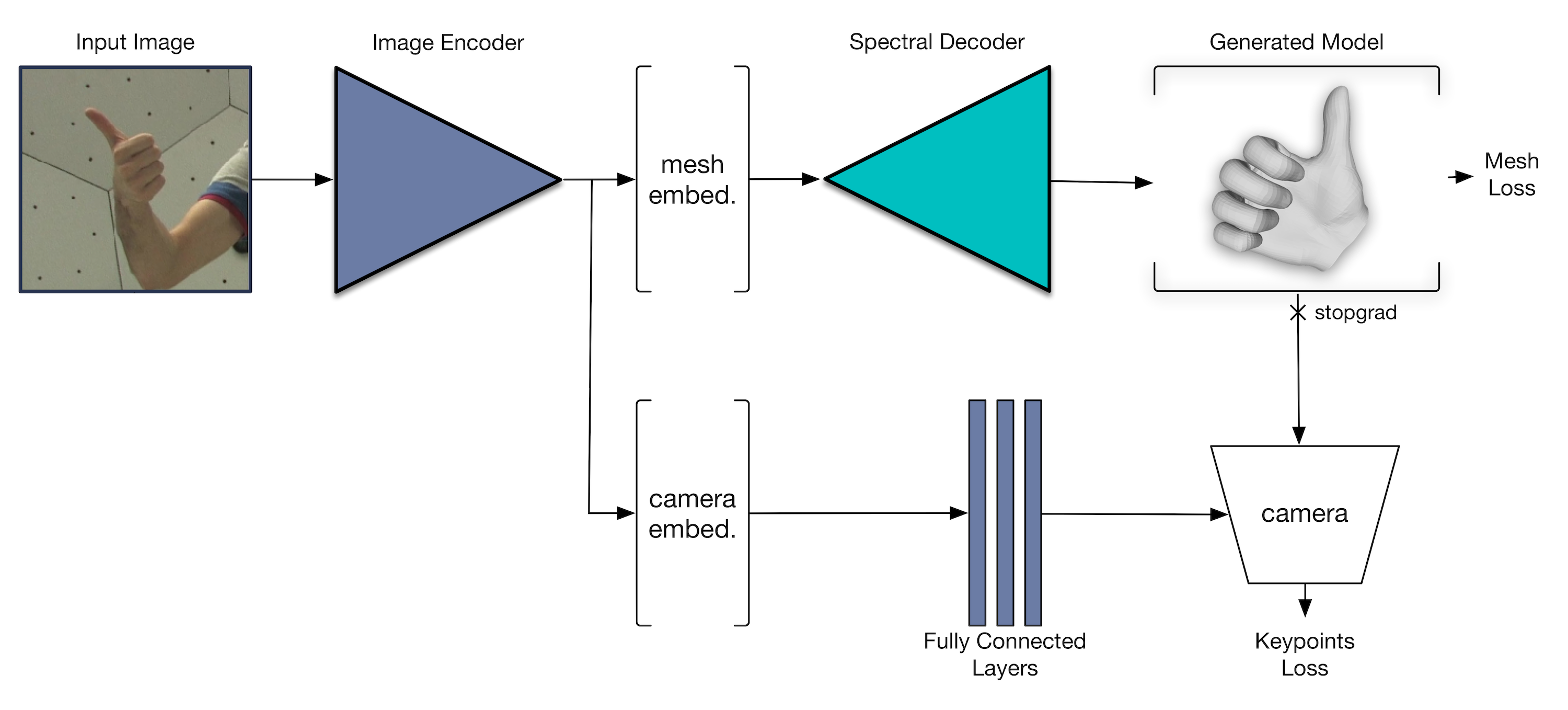}
\end{center}
   \caption{The presented system generates a 3D hand model corresponding to the hand in the image. The system consists of two branches trained simultaneously. The network in the top branch is capable of generating 3D hand models from a single image in a wide variety of shapes and poses. The ground truth meshes are aligned to the mean hand model in a canonical frame. The network in the lower branch regresses parameters of a weak perspective projection to align the generated mesh with the image. The gradient from the keypoints loss does not flow through the top branch to prevent the camera regressor from affecting the quality of mesh reconstructions.}
\label{fig:mesh-generator}
\end{figure*}

Convolutional Neural Networks (CNNs) have been effectively used in computer vision tasks on human geometry understanding, such as 3D model fitting and surface correspondence estimation.
Recent model based 3D reconstruction systems~\cite{hmr_kanazawa,DBLP:journals/corr/abs-1808-05942,2018arXiv180504092P} predict parameters of a statistical deformable model of the human body and a weak perspective camera for the alignment. These systems rely on the SMPL model~\cite{SMPL:2015}, where the shape of the person and deformations due to human pose are modelled with linear bases.
In this paper, we take a radically different approach and propose a new paradigm for 3D deformable alignment and reconstruction of articulated shapes by exploiting the intrinsic structure of the 3D hand meshes using graph convolutions. Our system is presented in Figure~\ref{fig:mesh-generator}.

\newpage
To summarise, our contributions are as follows: 
\begin{itemize}
    \item In order to create the high-quality training data, we build a new high-resolution model of the hand. We fit the model to around thousand scans to compute a distribution of valid pose and shape parameters and to 3D annotations from the Panoptic Studio dataset~\cite{DBLP:journals/corr/SimonJMS17} to obtain the ground truth data for the mesh recovery system.
    \item We train an autoencoder on sampled meshes with removed scale variations and global orientation. By reusing the decoder, we obtain the first graph morphable model of the human hand. Pose deformations are learned directly by a neural network and therefore the model does not suffer from computational drawbacks and training inefficiency of skinning methods with corrective offsets. It also has a significantly smaller number of parameters than the initial model and constraints the space of valid poses.
    \item We train an image encoder and take advantage of the pre-trained spectral decoder to recover 3D hand meshes (Figure~\ref{fig:mesh-generator}). The resulting system is able to generate hand models in real time. Additionally, we incorporate a camera regression network to compute a weak projection. We find that the network outcompetes the baseline method on the mesh reconstruction task.
\end{itemize}

Finally, we make all the contributed items (i.e. model, ground truth data, source code) publicly available at \href{https://github.com/dkulon/hand-reconstruction}{https://github.com/dkulon/hand-reconstruction}.

\section{Related Work} \label{related-work}

\paragraph{Statistical Modelling of 3D Shapes.}
The work that popularized statistical modelling of 3D shape and texture came from Blanz and Vetter~\cite{blanz1999morphable}. They constructed the first statistical 3D Morphable Model (3DMM) by performing dimensionality reduction on a set of meshes representing human faces with a fixed topological structure. It was a PCA-based implementation that produces new shape instances from a combination of linear basis. 

However, pose changes of the body joints and soft tissue deformations are non-linear and thus cannot be captured by PCA. The first deformable model of the human body was created by Allen et al.~\cite{Allen:2003:SHB:882262.882311}. They constructed a PCA model from 250 registered body scans in an A-pose coming from the CAESAR dataset. The pose changes are obtained by applying linear blend skinning which interpolates rotations matrices assigned to the joints to transform a vertex. This technique, widely used in computer graphics, causes the loss of volume of the surface close to the joints. 

The follow-up work~\cite{allen06learning} tries to solve this issue by adding a corrective offset. The Skinned Multi-Person Linear model (SMPL) \cite{SMPL:2015} uses a similar approach to produce more realistic results by learning pose-dependent corrective blend shapes from a large number of scans. MANO~\cite{MANO:SIGGRAPHASIA:2017} is an analogously defined hand model with 778 vertices introduced because SMPL does not include hand blend shapes. 

While SMPL-based models produce realistic results, they are heavyweight in terms of parameterization causing slow inference and the optimization process is time-consuming. Moreover, these models have a low number of vertices and it is arguable whether corrective offsets have enough representation power to capture fine-details of soft-tissue deformations. Nevertheless, they have become a standard tool for recovering human body parts where the problem is being solved by regressing axis-angles of joint rotations and blend shape weights given an image with keypoint annotations.

\paragraph{Body Mesh Recovery.} Numerous methods have been proposed to recover a subject-specific body mesh from an RGB image by finding parameters of the SMPL model. In SMPLify, Bogo et al.~\cite{bogo2016keep} estimate the 2D body joint locations and use these estimates to iteratively fit the SMPL model. The space of plausible shapes is constrained by adding prior and interpenetration error terms to the objective function. The shape parameters are only derived from the joint locations and thus the shape of the predicted model is close to the average. 

The issue of shape estimation inaccuracy is addressed in other works by introducing a silhouette error term. Lassner et al.~\cite{DBLP:journals/corr/Lassner0KBBG17} extend the SMPLify method by penalizing the bi-directional distance from points of the projected model to the ground-truth silhouette. This approach provides a better estimate of the body shape than its predecessor but it again requires numerically solving a complex optimization problem.

The Human Mesh Recovery framework \cite{hmr_kanazawa} reconstructs the body mesh in an end-to-end manner by regressing the model parameters and minimizing the reprojection loss of the joint locations given by the SMPL model and the ground-truth 2D image annotations. The authors introduce an adversary to discriminate body parameters that constrains the pose space.  Tan et al.~\cite{tan2017indirect} train a decoder from the SMPL parameters to a body silhouette and afterwards they train an encoder-decoder network that outputs a silhouette given an RGB image. Pavlakos et al.~\cite{2018arXiv180504092P} train a neural network that consists of modules responsible for predicting the mask, landmarks, and model parameters. Varol et al.~\cite{2018arXiv180404875V} combine segmentation, 2D pose, and 3D pose predictions to infer the volumetric representation of the body into which they fit the SMPL model. Neural Body Fitting~\cite{DBLP:journals/corr/abs-1808-05942} is an end-to-end system that integrates the SMPL model within a CNN. The network is a sequence of layers responsible for computing the segmentation, model parameters, and joint positions. Finally, the joints are projected onto the image to compute the loss. Weng at al.~\cite{DBLP:journals/corr/abs-1812-02246} show that it is possible to create an animated character from a single photo. 

All models described above have none or very limited variance of hand poses and face expressions. The authors of the Total Capture system \cite{DBLP:journals/corr/abs-1801-01615} build a unified body model that is iteratively fit into the point cloud obtained from a multi-camera setup. The system produces impressive results but is limited to the laboratory setting. The follow-up work \cite{DBLP:journals/corr/abs-1812-01598} enables predicting the model parameters in the wild by fitting the model into predicted joint confidence and orientation maps.

\paragraph{Hand Mesh Recovery.} Concurrently to our work, five methods have been proposed to recover a hand mesh from a single image. Four of them rely on regressing parameters of the MANO model whereas our hand model is represented as a lightweight generator trained end-to-end in an unsupervised manner.

The first work by Boukhayma et al.~\cite{2019arXiv190203451B} proposes a system analogous to the Human Mesh Recovery and is evaluated on images in the wild. The second work by Zhang et al.~\cite{2019arXiv190209305Z} introduces a method that iteratively regresses model parameters from the heatmaps. Baek et al.~\cite{2019arXiv190404196B} combine an iterative refinement with a differentiable renderer and Hasson et al.~\cite{Hasson:CVPR:2019} reconstruct hand meshes with objects they interact with. 

The last work by Ge et al.~\cite{2019arXiv190300812G} is similar to ours but uses depth maps for supervision and focuses on hand pose estimation while our goal is to recover a realistic hand model. All authors claim state-of-the-art performance on evaluated hand pose estimation benchmarks what shows the benefits of using a model-based representation.

\paragraph{Geometric Deep Learning}

Recent works in the area of geometric deep learning~\cite{bronstein2017geometric} generalized convolutions to graphs and Riemannian manifolds. Particularly, these methods allow us to extract features from a local neighbourhood of a vertex placed in a triangular mesh. 

Bruna et al.~\cite{DBLP:journals/corr/BrunaZSL13} expressed convolutions in the spectral domain where the filter is parameterized using a smooth spectral transfer function. This method has multiple disadvantages. The filters are basis-dependent and thus do not generalize across domains. The method requires $O(n)$ parameters per layers and the computation of forward and inverse Fourier transforms has $O(n^2)$ complexity. Moreover, there is no guarantee of spatial localization of filters. 

Defferrard et al.~\cite{DBLP:journals/corr/DefferrardBV16} represent a spectral transfer function as a polynomial of degree $r$. This approach has a constant number of parameters per layer and filters are guaranteed to have $r$-hops support. There is also no explicit computation of the eigenvectors resulting in $O(nr)$ computational complexity. CayleyNet~\cite{DBLP:journals/corr/LevieMBB17} replaced polynomial filters with rational complex functions to avoid the Laplacian eigendecomposition and achieve better spectral resolution. 

The first mesh convolutional neural networks such as Geodesic CNN~\cite{7406461}, Anisotropic CNN~\cite{DBLP:journals/corr/BoscainiMRB16}, and MoNet~\cite{DBLP:journals/corr/MontiBMRSB16} define convolutions in the spatial domain where it is possible to orient the kernels. FeastNet~\cite{DBLP:journals/corr/VermaBV17} uses an attention mechanism to weight the neighbour selection whereas spiral filters traverse adjacent vertices in a fixed order \cite{DBLP:journals/corr/abs-1809-06664}. 

Ranjan et al.~\cite{DBLP:journals/corr/abs-1807-10267} use fast spectral convolutions~\cite{DBLP:journals/corr/DefferrardBV16} to find a low-dimensional non-linear representation of the human face in an unsupervised manner. Their network has 75\% less parameters than a PCA-based morphable model while obtaining improved reconstruction results. Cheng et al.~\cite{DBLP:journals/corr/abs-1903-10384} define this architecture in an adversarial setting obtaining more detailed reconstructions and Zhou et al.~\cite{zhou2019dense} include additional features in the graph structure to predict the face shape with texture. In this work, we show for the first time that mesh autoencoders, if appropriately trained, can be used to represent highly-articulated objects such as hands.  

\section{Graph Morphable Model}

We introduce the first hand model learned form a collection of meshes in an unsupervised way. The model is capable of generating realistic posed hand shapes with 7,907 vertices. We achieve this by training a mesh autoencoder with convolutions performed in the frequency domain. The decoder has only four layers resulting in low computational requirements. Interpolation in the latent space produces meaningful transitions enabling fast optimization of the objective function for the problem of model fitting by regressing latent parameters (Figure~\ref{fig:mesh-generator}).

\subsection{Graph Fourier Transform}

Non-normalized graph Laplacian is defined as $\Delta = D - W$ where $D = diag(\sum_{j}W_{ij})$ is the degree matrix and $W$ a weighted adjacency matrix. For any signal $\vec f: V \rightarrow \!R^N$ defined on an undirected graph $G = (V, \mathcal{E}, W)$, the graph Laplacian satisfies 
\begin{equation}
(\Delta f)(i) = \sum_{j \in N_i} W_{ij}(f(i) - f(j))
\end{equation}
where $N_i = \{j\, |\, (i, j) \in \mathcal{E}\}$ it the neighbourhood of a vertex $i$, $V$ a set of vertices, and $\mathcal{E}$ a set of edges \cite{Chung:1997, DBLP:journals/corr/abs-1211-0053}. 

Graph Laplacians are symmetric and positive semi-definite. Therefore they have a complete set of orthogonal eigenvectors $\Phi = (\phi_1, ..., \phi_n)$ where $n = |V|$ with real, non-negative eigenvalues $\lambda_1, ..., \lambda_n$. 
Due to these properties, $\Delta$ admits an eigendecomposition $\Delta = \phi \Lambda \phi^{T}$ where $\Lambda = (\lambda_1, ..., \lambda_n)$. Graph Fourier transform is the expansion of $\vec f$ in terms of the eigenvectors of the graph Laplacian which can be written in the matrix form $\hat{\vec f} = \phi^T \vec f$. It follows that the inverse graph Fourier transform is given by  $\vec f = \phi \hat{\vec f}$.

\subsection{Spectral Convolutions}

Bruna et al.~\cite{DBLP:journals/corr/BrunaZSL13} expressed convolution in Fourier space $(f \star h) = \hat{h}(\Delta)f$ where the filter $\hat{h}$ is parameterized using a smooth spectral transfer function
\begin{align}
    \begin{split}
        \hat{h}(\Delta) = \Phi \hat{h}(\Lambda) \Phi^T \\
    \end{split} \\
    \begin{split}
        \hat{h}(\Lambda) = diag(\hat{h}(\lambda_1), ..., \hat{h}(\lambda_n)).
    \end{split}
\end{align}
Defferrard et al.~\cite{DBLP:journals/corr/DefferrardBV16} parameterized the filter with a Chebyshev expansion of order $r - 1$ such that 
\begin{equation}
    h(\tilde{\Delta}) = \sum_{j=0}^{r-1} \alpha_j T_j(\tilde{\Delta})
\end{equation}
where $T_k(\lambda) = 2 \lambda T_{k-1}(\lambda) - T_{k-2}(\lambda)$ with $T_0 = 1$ and $T_1 = \lambda$. $\tilde{\Delta} = 2 \lambda^{-1}_{n}\Delta - \mathrm{I}$ denotes rescaling of the Laplacian eigenvalues from the interval $[0, \, \lambda_n]$ to $[-1, 1]$. As we discussed in the Related Work section, this approach is computationally faster and filters are localized with $r$-hops support.

\subsection{Network Architecture}

We follow design choices of CoMA~\cite{DBLP:journals/corr/abs-1807-10267} which is a mesh autoencoder with four layers of convolutions followed by downsampling. We start with input mesh with 7,907 vertices followed by sequence of convolutions (16, 32, 32, 48 filters) and downsampling after each convolutional layer (4, 4, 2, 2 graph reduction factors). Afterwards, we apply a fully connected layer to obtain a latent vector with 64 parameters. The decoder is symmetric. We choose leaky ReLU for the activation function based on experimental evaluation and we use a smaller filter with the Chebyshev polynomial of order $r=3$. Downsampling approach is also adopted from CoMA which minimises the quadric error to decimate the template. However, we compute downsampled topology from a mesh in a half-closed hand position in Blender. We find that the network trained with the original implementation produces extremely noisy meshes because decimated templates do not have vertices around the joints. The choice of downsampled graph topology has the most significant effect on the quality of mesh reconstructions. 

The training data comes from a MANO-like model with 7,907 vertices. The process of sampling realistic templates is described in Section~\ref{data-acq}. We train the network for 6 hours on a single GeForce RTX 2080 Ti with a batch size 64 and learning rate 0.001. In addition to the L1 reconstruction loss, we impose an L2 penalty on the latent vector weighted by 5e-7 and we use L2 regularization scaled by 5e-5. The loss function is minimized with AdamW opimizer with a decay factor 10e-6. Our Graph Morphable Model (GMM) is the decoder with $Z=64$ latent parameters.

\section{Single Image Mesh Generation}

We take the dataset of hand images with corresponding 3D hand meshes aligned in canonical coordinates (Section~\ref{data-acq}). The single image mesh generator (Figure~\ref{fig:mesh-generator}) consists of the image encoder ${E}_{image}$, for which we use the DenseNet-121~\cite{DBLP:journals/corr/HuangLW16a} network, pretrained on the ImageNet classification task. The outputs of the ${E}_{image}$ are a mesh embedding that is passed to the pre-trained Graph Morphable Model $\mathcal{D}_{GMM}$, and a camera embedding fed into fully connected layers to estimate parameters of a weak perspective camera $\mathcal{D}_{camera}$. The hand joints, obtained by taking the average of the surrounding ring vertices in the generated mesh, are projected onto the image plane to compute the loss based on 2D keypoint annotations. More specifically, we minimize the loss:
\begin{equation}
\begin{split}
\mathcal{L} & =
 \sum_i|\hat{\mathcal{Y}}_{i} - \mathcal{D}_{GMM}({E}_{image}(X_i))|_1 \\ 
 & + \lambda_{kpts} \sum_i|\hat{\mathcal{J}}_{i} - \mathcal{D}_{camera}({E}_{image}(X_i))|_1 \\
& +  \lambda_{embed} \sum_i||{E}_{image}(X_i)_{1:Z}||_2
\end{split}
\end{equation}
for ground truth meshes $\hat{\mathcal{Y}}_{i}$, keypoint annotations $\hat{\mathcal{J}}_{i}$, and input images $X_i$. We set hyperparameters $\lambda_{kpts}$ to 0.01 and $\lambda_{embed}$ to 5e-5. We also add L2 regularization weighted by 1e-5. 

During training, we freeze the weights of the GMM and train the image encoder and camera regressor simultaneously for 130 epochs with the same hardware setting and optimizer as the autoencoder but we set learning rate to 1e-4. The network is able to reproduce hand pose early in the training while longer optimization reduces noise around fingertips for extreme poses. We zero the gradient that flows from the camera regression module through $\mathcal{D}_{GMM}$ as we find it to provide better reconstruction results. 

\section{Evaluation} \label{data-acq}

There are no existing benchmarks that contain images of hands with corresponding meshes or large collection of meshes that can be used to train the Graph Morphable Model and a body part recovery system. Therefore, to address both issues we build a new high resolution hand model following MANO with ten times more vertices and removed scale variations from PCA linear bases. Afterwards, we fit this model to 3D keypoints annotations from the Panoptic dome dataset~\cite{DBLP:journals/corr/SimonJMS17}. We also compute a distribution of valid poses from registrations of around a thousand scans from the MANO dataset to sample realistic meshes for training the autoencoder. 
\paragraph{Panoptic DomeDb.} This dataset provides 3D annotations of the whole human body including 21 hand joints. It contains multiple sequences of videos with a large number of subjects captured from multiple views in a laboratory setting. We use a subset of around 30 synchronized HD cameras and sample best annotations from a single camera in terms of visibility after each 100th frame. Our preprocessed dataset contains 40360 training, 3000 validation, and 3000 test samples with 3D keypoint annotations projected to the image coordinates. 

\begin{figure*}
     \centering
     \includegraphics[width=0.98\linewidth]{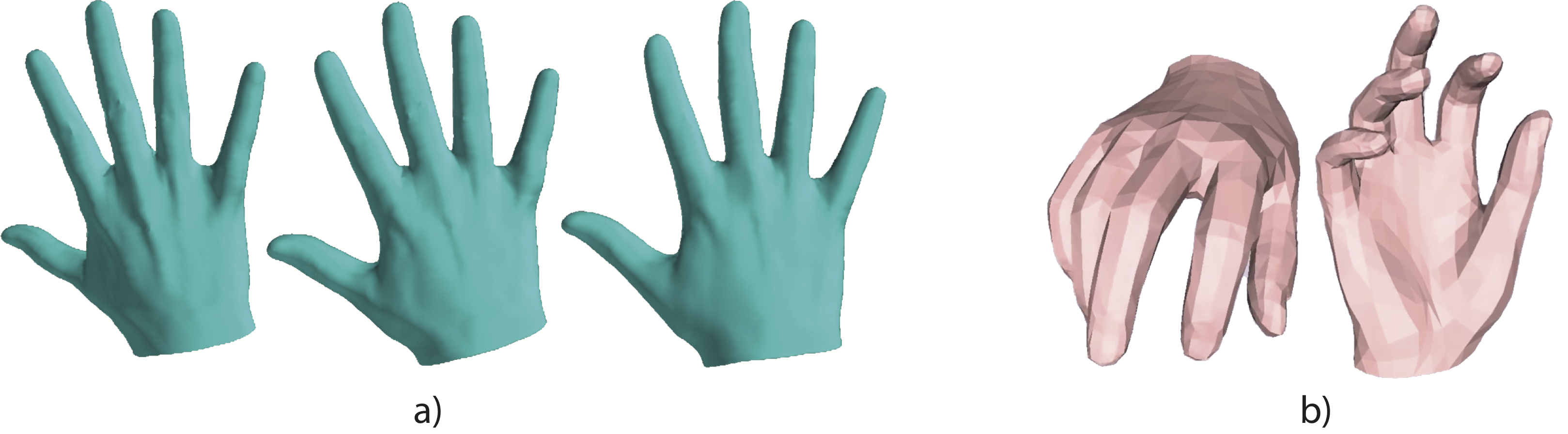}
    \caption{ a) Random shapes sampled from our high resolution linear model. Posed examples can be found in Figure~\ref{fig:results-quality}. b) Samples from the MANO model (taken from \cite{MANO:SIGGRAPHASIA:2017}). }
    \label{Fig:pca-samples}
\end{figure*}

\paragraph{Hand Model.} The number of vertices in MANO (778) is not sufficient to realistically model shape deformations of the human hand. A potential increase in the number of vertices introduces difficulties with scan registrations and significantly increases optimization time for the tasks of parameter training and pose fitting. Moreover, the shape model in MANO was created by applying a dimensionality reduction to rigidly aligned scans. This results in a first principal component that models scale deformations. This is undesirable for our model fitting procedure that applies a similarity alignment at the beginning of optimization. To address these issues, we introduce a hand model with 7,907 vertices which is more than SMPL and MANO combined. We register the reference template to a set of scans with selected keypoints. Then the shape components are computed by applying PCA to the set of Procrustes aligned registrations. The size of the model is controlled by a scale multiplier. Figure~\ref{Fig:pca-samples} shows that we are able to model fine-details of the hand surface. Finally, we optimize pose, shape, and perspective parameters with the Dogleg method to match the ground truth keypoint annotations~\cite{bogo2016keep}.

\paragraph{Pose Sampler.} We use samples obtained from our high-resolution linear model to train the proposed mesh autoencoder. The shape coefficients are sampled from a standard normal distribution. In order to sample plausible and diverse hand poses, we resort to the distribution of joint angles in the MANO database. For each 15 joint angles in the human hand kinematic tree, we compute euler-angle clusters via K-means~\cite{Guler_2019_CVPR}. During synthesis of training samples, we randomly select euler-angle cluster centers for each joint, effectively sampling shapes from the manifold of plausible shapes. In our experiments, we have used 64 rotation centers.

\begin{figure}[t]
\begin{center}
  \includegraphics[width=1\linewidth]{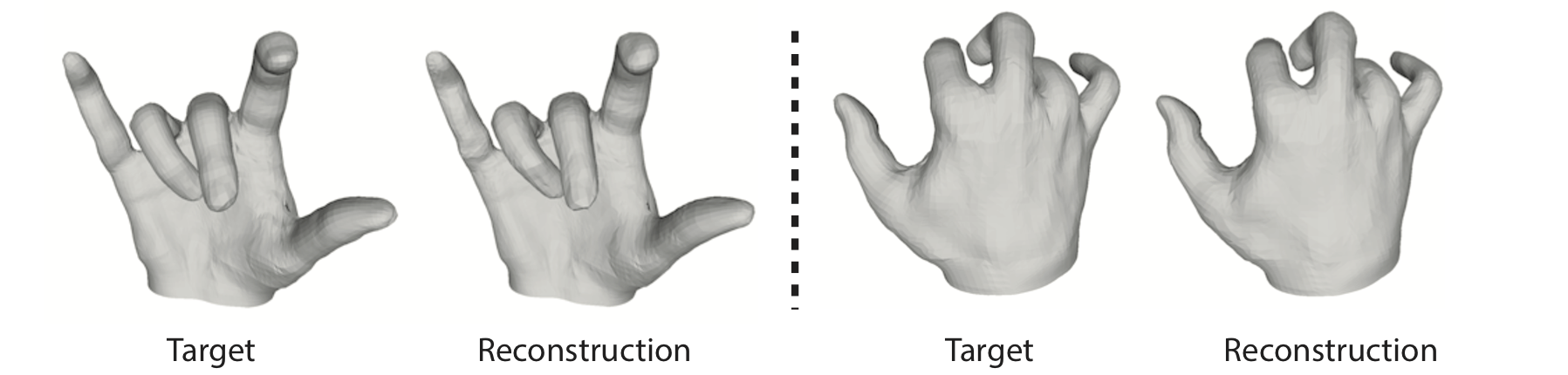}
  \caption{Autoencoder reconstruction results for challenging samples in terms of the pose (left pair) and shape (right pair).}
  \label{fig:ae-results}
  \end{center}
\end{figure}

\begin{figure}
\begin{center}
  \includegraphics[width=0.98\linewidth]{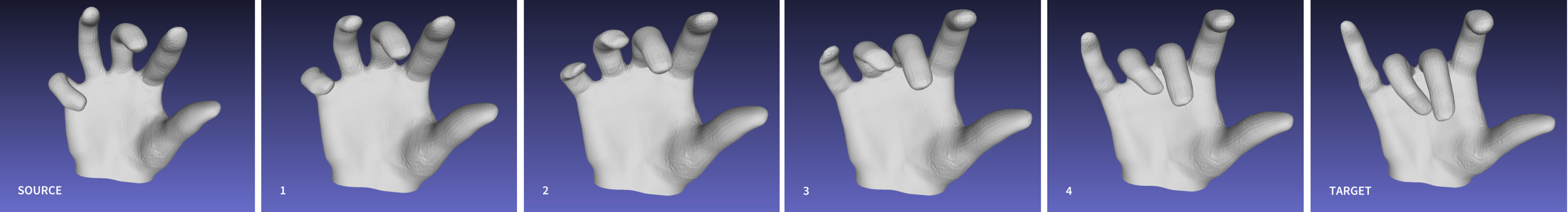}
  \caption{Interpolation in the latent space between two random samples. Please, note that source and target shapes (e.g. finger length) are also different.}
  \label{fig:interpolation}
  \end{center}
\end{figure}

\begin{figure}[H]
\begin{center}
  \includegraphics[width=0.98\linewidth]{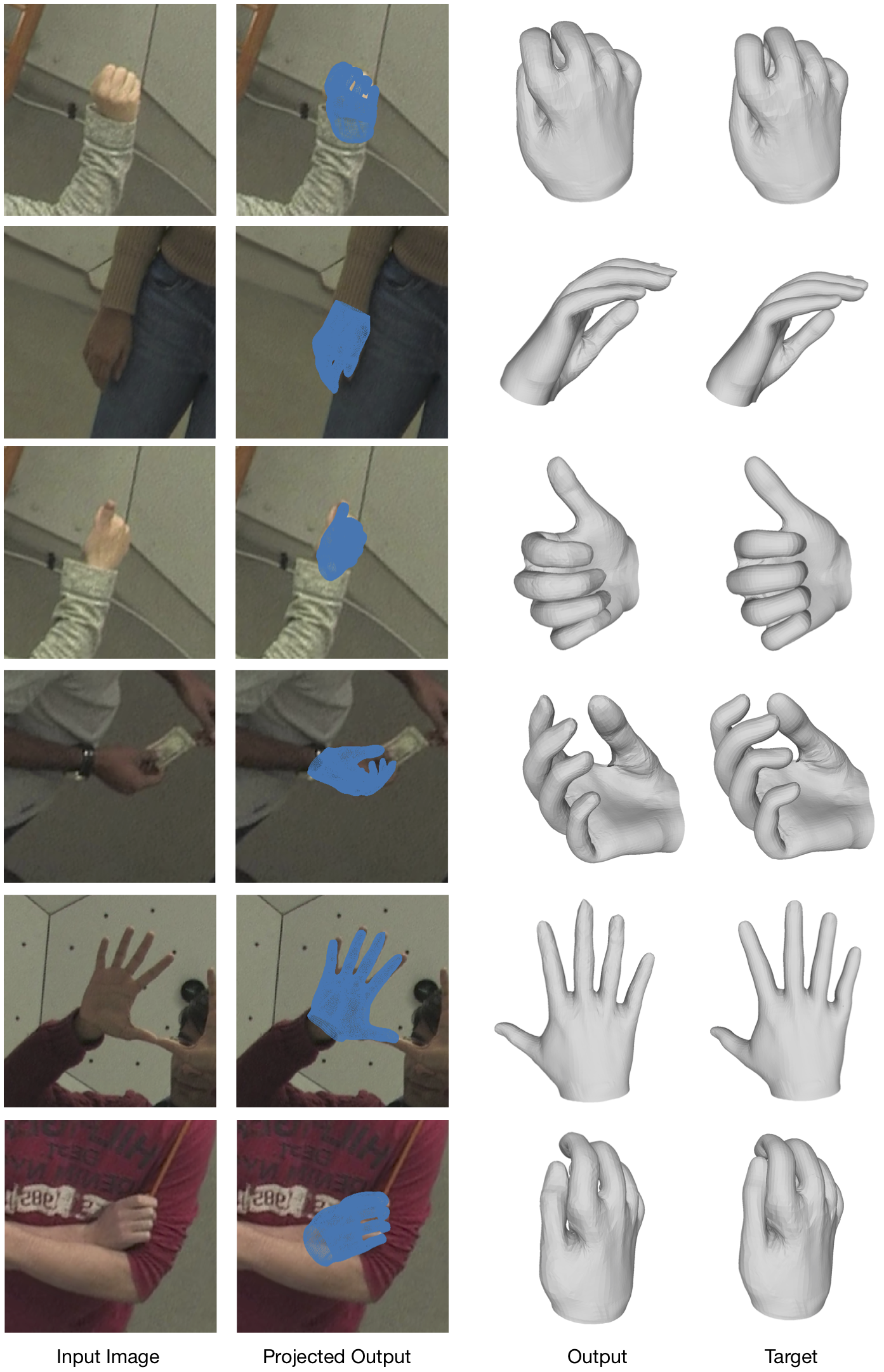}
    \caption{Qualitative results of our system.}
    \label{fig:results-quality}
\end{center}
\end{figure}

\paragraph{Results.} We implement a baseline method that replaces a spectral decoder from Figure~\ref{fig:mesh-generator} with a TensorFlow implementation of our high resolution MANO model (MANO-like). We also fine-tune the network (Spectral, fine-tuned) by training our network (Spectral, fixed) for 170 more iterations without freezing the decoder's weights. Table~\ref{table:panoptic-eval} shows that we are able to obtain a lower reconstruction error despite the fact that the MANO model was used to generate the ground truth data in the first place. The table also presents that the spectral decoder is six times smaller in terms of number of parameters, obtains lower inference time, and operates in real-time. The DenseNet-121 image encoder runs at 53 FPS, therefore, the whole system maintains real-time performance. Qualitatively, we show that the autoencoder produces visually indistinguishable reconstructions from the input (Figure~\ref{fig:ae-results}) and has a meaningful latent representation providing smooth transitions in the latent space (Figure~\ref{fig:interpolation}). The visualization of the results of the mesh recovery system is presented in Figure~\ref{fig:results-quality}.

\begin{table}[t]
\begin{center}
\begin{tabular}{l|c|c|c|}
\cline{2-4}
& Spectral, fixed & Spectral, fine-tuned                                          & MANO-like                           \\ \hline
\multicolumn{1}{|l|}{Reconstruction error {[}mm{]}}  & 2.33            & \textbf{2.30}  & 2.56           \\ \hline
\multicolumn{1}{|l|}{Inference time (generator) {[}ms{]}}        & -           & \textbf{3.04} & 4.64         \\ \hline
\multicolumn{1}{|l|}{Inference time (generator) {[}fps{]}}       & -             & \textbf{ 329}                           & 216            \\ \hline
\multicolumn{1}{|l|}{Number of params. (generator)}       & \textbf{393,080}              & \textbf{393,080}                           &      2,498,612      \\ \hline
\end{tabular}
\end{center}
\caption{Mesh L1 reconstruction error, inference time, and number of parameters for systems with different types of generators evaluated on Panoptic DomeDb. The scale of the target meshes is smaller than in real world.}
\label{table:panoptic-eval}
\end{table}

\newpage

\section{Conclusion}

We proposed a system for generating a subject-specific hand model from a single image. To achieve this, we trained a graph morphable model in an unsupervised manner obtaining a lightweight non-linear representation of hand shapes. The resulting generator was connected to the image encoder and camera regression networks to produce meshes aligned with images. Our system is able to produce realistic hand shapes in real-time that match the target models. To train the networks, we generated a dataset of images with corresponding 3D meshes using a high-vertex count hand model with blend shapes learned from scans. 

The morphable model could benefit from defining kernels in a different domain. Spectral methods do not impose canonical ordering of neighbours because the kernels are isotropic due to rotation invariance of the graph Laplacian. In the spatial domain, we can address the issue of orientation ambiguity by imposing a canonical direction or angular max pooling. We expect that spatial localization of filters would allow the network to model fine-details of body part deformations in contrast to spectral methods that average the neighbours \cite{2019arXiv190502876B}. We observed that the choice of a downsampled graph structure has the utmost importance on the performance of the network. Therefore, the need for a learnable graph coarsening mechanism \cite{2018arXiv180608804Y, 2018arXiv181101287C} naturally arises in our scenario. We will explore spatial mesh convolutions and differentiable pooling in the future works. In terms of applications beyond computer graphics, our system can be modified to solve the problems of hand pose estimation \cite{DBLP:journals/corr/SimonJMS17, DBLP:journals/corr/ZimmermannB17, spurr2018cvpr}, fingertip detection \cite{WetzlerBMVC15}, or dense correspondence computation \cite{DBLP:journals/corr/GulerTASZK16, Guler2018DensePose}.

\section{Acknowledgements} The work of S. Zafeiriou and R. A. G\"uler has been partially funded by the EPSRC Fellowship Deform (EP/S010203/1). The work of M. Bronstein has been partially funded by ERC Consolidator Grant No. 724228 (LEMAN). Finally, the work of D. Kulon was funded by a PhD scholarship. 
\newpage

\newpage

\bibliography{egbib}
\end{document}